\documentclass[10pt,twocolumn,letterpaper]{article}

\usepackage{cvpr}
\usepackage{times}
\usepackage{epsfig}
\usepackage{graphicx}
\usepackage{amsmath}
\usepackage{amssymb}
\usepackage{mathtools}
\usepackage{graphicx}
\usepackage{caption}
\usepackage{subcaption}
\usepackage{stackengine} 
\usepackage{mathbbol}

\usepackage[pagebackref=true,breaklinks=true,letterpaper=true,colorlinks,bookmarks=false]{hyperref}

\cvprfinalcopy 

\newcommand \enc {\text{Enc}}
\newcommand \dec {\text{Dec}}
\newcommand \dsc {\text{Dsc}}
\newcommand \cls {\text{Cls}}

\newcommand \m {\mathbf{m}}
\newcommand \x {\mathbf{x}}
\newcommand \y {\mathbf{y}}
\newcommand \f {\mathbf{f}}
\newcommand \w {\mathbf{w}}

\newcommand \celebasize {0.24}
\newcommand \spritessize {0.235}

\def\cvprPaperID{2797} 

\ifcvprfinal\fi
\begin{document}

\title{Disentangling Factors of Variation by Mixing Them}

\author{Qiyang Hu\textsuperscript{1}\thanks{The authors contributed equally.} ~~
Attila Szab\'{o}\textsuperscript{1}\footnotemark[1] ~~
Tiziano Portenier\textsuperscript{1} ~~
Paolo Favaro\textsuperscript{1} \\
\textsuperscript{1}University of Bern, Switzerland \\
{\tt\small \textsuperscript{1}\{hu, szabo, portenier, favaro\}@inf.unibe.ch} \\
\and Matthias Zwicker\textsuperscript{2}\\
\textsuperscript{2}University of Maryland, USA \\
{\tt\small \textsuperscript{2}zwicker@cs.umd.edu}
}

\maketitle

\begin{abstract}
We propose an approach to learn image representations that consist of disentangled factors of variation without exploiting any manual labeling or data domain knowledge. A factor of variation corresponds to an image attribute that can be discerned consistently across a set of images, such as the pose or color of objects. Our disentangled representation consists of a concatenation of feature chunks, each chunk representing a factor of variation. It supports applications such as transferring attributes from one image to another, by simply mixing and unmixing feature chunks, and classification or retrieval based on one or several attributes, by considering a user-specified subset of feature chunks. We learn our representation without any labeling or knowledge of the data domain, using an autoencoder architecture with two novel training objectives: first, we propose an invariance objective to encourage that encoding of each attribute, and decoding of each chunk, are invariant to changes in other attributes and chunks, respectively;
second, we include a classification objective, which ensures that each chunk corresponds to a consistently discernible attribute in the represented image, hence avoiding degenerate feature mappings where some chunks are completely ignored. We demonstrate the effectiveness of our approach on the MNIST, Sprites, and CelebA datasets.
%
%
   
\end{abstract}

\section{Introduction}

Deep learning techniques have led to highly successful natural image representations, some focusing on synthesis of detailed, high resolution images of photographic quality~\cite{chen2017photographic,karras2017progressive}, and others on disentangling image features into semantically meaningful properties~\cite{Chen2016InfoGAN,mathieu2016disentangling,reed2015deep}. 

In this paper, we learn a disentangled image representation that separates the feature vector into multiple chunks, each chunk representing intuitively interpretable properties, or factors of variation, of the image. We propose a completely unsupervised approach that does not require any labeled data, such as pairs of images where only one factor of variation changes (different viewpoints, for example)~\cite{mathieu2016disentangling, szabo2017challenges}.
The basic assumption of our technique is that images can be represented by a set of factors of variation, each one corresponding to a semantically meaningful image attribute. In addition, each factor of variation can be encoded using its own feature vector, which we call a feature chunk. That is, images are simply represented as concatenations of feature chunks, in a given order. We obtain disentanglement of feature chunks by leveraging autoencoders, and as a key contribution of this paper, by developing a novel invariance objective. The goal of the invariance objective is that each attribute is encoded into a chunk invariant to changes in other attributes, and that each chunk is decoded into an attribute invariant to changes in other chunks. We implement this objective using a sequence of two feature mixing and unmixing autoencoders.


The invariance objective using feature mixing on its own, however, does not guarantee that each feature chunk represents a meaningful 
factor of variation. Instead, the autoencoder could represent the image with a single chunk, and ignore all the others. This is called the shortcut problem~\cite{szabo2017challenges}. 
We address the shortcut problem with a classification constraint, which forces each chunk to have a consistent, discernible effect on the generated image.

We demonstrate successful results of our approach on several datasets, where we obtain representations consisting of feature chunks that determine semantically meaningful image properties. In summary, we make the following contributions:
\emph{1)} A novel architecture to learn image representations of disentangled factors of variation without using any annotation or data domain knowledge, and where the representation consists of a concatenation of a fixed number of feature chunks. Our approach can learn several factors of variation simultaneously; \emph{2)} A novel invariance objective to obtain disentanglement by encouraging invariant encoding and decoding of image attributes and feature chunks, respectively;
\emph{3)} A novel classification constraint to ensure that each feature chunk represents a consistent, discernible factor of variation of the represented image;
\emph{4)} An evaluation on the MNIST, Sprites, and CelebA datasets to demonstrate the effectiveness of our approach.


\section{Related work}

\noindent\textbf{Autoencoders.}
Our architecture is built on autoencoders~\cite{bourlard1988auto, hinton2006reducing, bengio2013representation}, which are neural networks with two main components: an encoder and a decoder. The encoder is designed to extract a feature representation of the input (image), and the decoder translates the features back to the input.
Different flavors of autoencoders have been trained to perform image restoration \cite{vincent2008extracting,mao2016image,bigdeli2017meanshift} or image transformation \cite{hinton2011transforming}. While basic autoencoders do not impose any constraints on the representation itself, variational autoencoders \cite{kingma2013auto} add a generative probabilistic formulation, which forces the representation to follow a Gaussian distribution and allows sampling images by applying the decoder to a Gaussian vector sample.
Thanks to their flexibility, autoencoders have become ubiquitous tools in large systems for domain adaptation \cite{CycleGAN2017, kim2017learning}, or unsupervised feature learning \cite{pathakCVPR16context}.  Autoencoders are also used to learn feature disentangling \cite{reed2015deep, mathieu2016disentangling, szabo2017challenges}. In our work we also use them as feature extractors. Our contribution is a novel unsupervised training method that ensures the separation of factors of variation into several feature chunks.
~\\\noindent\textbf{GANs.}
Generative Adversarial Networks (GANs) \cite{goodfellow2014generative} are designed to provide samples from a data distribution specified as a finite set of real data samples. They use two competing neural networks: a generator translates input noise vectors into fake data samples, while a discriminator tries to distinguish fake samples from real ones. In the ideal case, the trained generator produces convincing data samples of the real data distribution, and the trained discriminator cannot tell them apart from real ones. 
GANs have been successful at image to image translation \cite{pix2pix2016}, learning representation \cite{dcgan2015}, sampling images from a specific domain \cite{CycleGAN2017}, or ensuring that image-feature pairs have the same distribution when computing one from another \cite{donahue2016adversarial}. As the adversarial loss constrains the distribution of the generated data but not the individual data samples, it allows to reduce the need for data labeling. In particular, Shrivastava~\etal~\cite{Shrivastava_2017_CVPR} use GANs to transfer known attributes of synthetic, rendered examples to the domain of real images, thus creating virtually unlimited datasets for supervised training.
In our work we use GANs to enforce that images look realistic when their attributes are transferred.
~\\\noindent\textbf{Disentangling.}
There are many methods \cite{Tran_2017_CVPR, peng2017invariantface} that disentangle factors of variation by using manual annotation.
Kulkarni \etal~\cite{kulkarni2015deep} sample the data during the training, such that only one factor changes within a minibatch. They associate a feature chunk to the variation of the images in the minibatch.
One of the most immediate methods for disentangling is to mix the feature encodings of two input images with common known attributes in an autoencoder \cite{reed2015deep} and then train a decoder to map the mixed features to the ground truth image with mixed attributes.
In other methods, GANs and adversarial training have been leveraged to reduce the need for complete labeling of all factors of variation. 
For example, Mathieu \etal~\cite{mathieu2016disentangling} apply adversarial training on the image domain, while Denton \etal~\cite{denton2017unsupervised} propose adversarial training on the feature domain.
Szab{\'o} \etal~\cite{szabo2017challenges} studied the ambiguities in weakly supervised disentanglement. They can provably avoid a degenerate solution called the shortcut problem, where the complete image representation is condensed in only one feature chunk.


In some approaches, the physics of the image formation model is integrated into the network training, with factors like the depth and camera pose \cite{zhou2017unsupervised} or the albedo, surface normals and shading \cite{NeuralFace2017}. Shu \etal~\cite{NeuralFace2017} do no use any label from the training data. However, an externally trained 3D morphable model guides the training, which is also a form of annotation.

By maximizing the mutual information between synthesized images and latent features, InfoGAN \cite{Chen2016InfoGAN} makes the latent features interpretable as semantically meaningful attributes. InfoGAN is completely unsupervised, but it does not include an encoding stage. In contrast, we build on an autoencoder, which allows us to recover the disentangled representation from input images, and swap attributes between them. In addition, we use a novel classification constraint instead of the feature consistency in InfoGAN.

Two recent techniques, $\beta$-VAE \cite{higgins2016beta} and DIP-VAE \cite{dipvae2017}, build on variational autoencoders (VAEs) to disentangle interpretable factors in an unsupervised way, similarly to our approach. They encourage the latent features to be independent by generalizing the KL-divergence term in the VAE objective, which measures the similarity between the prior and posterior distribution of the latent factors. Instead, we build on mixing autoencoders~\cite{reed2015deep} and adversarial training~\cite{goodfellow2014generative}. We encourage disentanglement using an invariance objective, rather than trying to match an isotropic Gaussian prior.
Notice that our feature space is only designed for attribute transfer and not for sampling. Finally, we can use high-dimensional feature chunks, while in \cite{higgins2016beta} and \cite{dipvae2017} the chunks are one-dimensional.

\section{Unsupervised Disentanglement of Factors of Variation}

A representation of images where the factors of variations are disentangled can be exploited for various computer vision tasks. At the image level, it allows to transfer attributes from one image to another. At the feature level, this representation can be used for image retrieval and classification. To achieve this representation and to enable the applications at both the image and feature level, we leverage autoencoders. Here, an encoder transforms the input image $\x$ to its feature representation $\f = \enc(\x)$, where $\f = [f^1, f^2, \hdots, f^n]$ consists of multiple chunks $f^i \in \mathbf{R}^d$. The dimension of the full feature is therefore $n \times d$. In addition, a decoder transforms the feature representation back to the image via $\dec(\f) = \x$. 

Our main objective is to learn a disentangled representation, where each feature chunk corresponds to an image attribute. For example, when the data $\x$ are face images, chunk $f^1$ could represent the hair color, $f^2$ the gender and so on. With a disentangled representation, we can transfer attributes from one image to another simply by swapping the feature chunks.
An image $\x_3 = \dec( [f_1^1, f_2^2, f_2^3, \hdots, f_2^n] )$ could take the hair color from image $\x_1$ and all the other attributes from $\x_2$.

In our approach, we interpret disentanglement as invariance. In a disentangled representation, the encoding of each image attribute into its feature chunk should be invariant to transformations of any other image property. Vice versa, the decoding of each chunk into its corresponding attribute should be invariant to changes of other chunks. In our example, if $\x_1$ and $\x_2$ have the same gender, we must have $f_1^2 = f_2^2$ irrespective of any other attribute. Hence, a disentangled representation is also useful for image retrieval, where we can search for  nearest neighbors of a specified attribute. Invariance is also beneficial for classification, where a simple linear classifier is sufficient to classify each attribute based on its corresponding feature chunk.
This observation inspired previous work~\cite{dipvae2017} to quantify disentanglement performance using linear classifiers on the full features $\f$.

In the following, we describe how we learn a disentangled representation from data without any additional knowledge (\eg, labels, data domain) by using mixing autoencoders. 
One of the main challenges in the design of the autoencoder and its training is that the encoder and the decoder could just make use of a single feature chunk (provided that this is sufficient to represent the whole input image) and ignore the other chunks. We call this failure mode a \emph{shortcut} taken by the autoencoder during training.
We propose a novel invariance objective to obtain disentanglement, and a classification objective to avoid the shortcut problem.




\subsection{Network Architecture}

\begin{figure}[t]
	\begin{center}
	\includegraphics[width=.9\linewidth,trim=0 .9cm 0 1.7cm,clip]{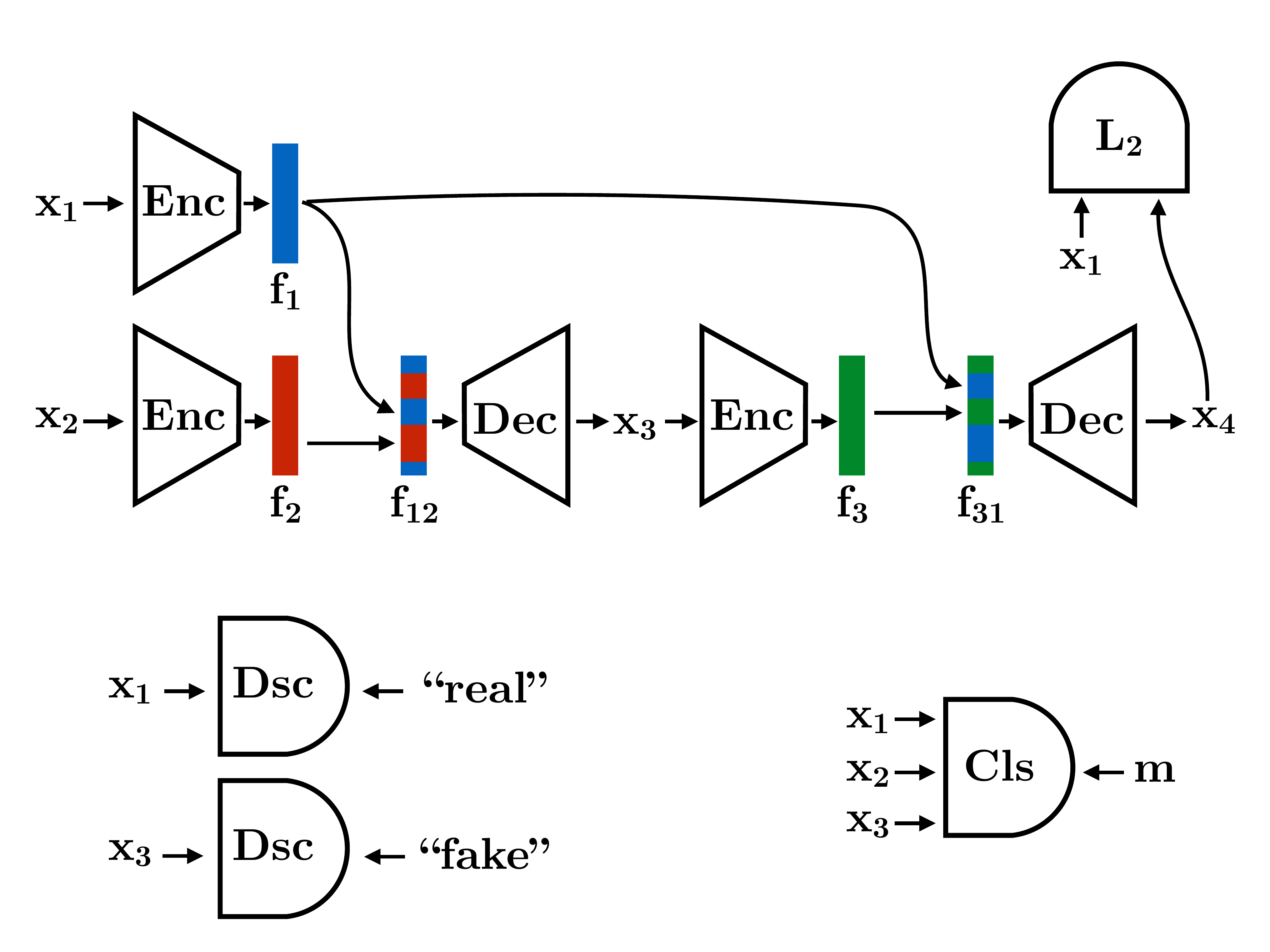}
	\end{center}
	\vspace{-.5cm}
	\caption{Overview of our architecture. The core component is a sequence of two mixing autoencoders (top). This implements our invariance objective, which encourages that the decoding of each feature chunk into an image attribute is invariant to a perturbation (mixing) in other chunks, and similarly, the encoding of each attribute into a chunk is invariant to a perturbation of other attributes. We include an adversarial loss to ensure the intermediate images obtained by perturbing some chunks is from our data distribution (bottom left). Finally, a classification objective avoids the shortcut problem, where chunks would be ignored completely. Components with the same name share weights.}
	\label{fig:architecture}
\end{figure}

Our network architecture is shown in Figure~\ref{fig:architecture}. There are three main components: We enforce invariance using a \emph{sequence of two mixing autoencoders}, and a \emph{discriminator}; we avoid the shortcut problem using a \emph{classifier}. They are all implemented as neural networks.

\noindent\textbf{Mixing/Unmixing Autoencoders.} 
We leverage a sequence of two mixing autoencoders to enforce invariance, ensuring that
we encode each attribute into a feature chunk invariant to changes in other attributes, and that we decode each chunk similarly in an invariant manner into its attribute. 
%
%
More precisely, the sequence of two mixing autoencoders performs the following operations (Figure~\ref{fig:architecture}):
\begin{enumerate}
\item Sample two images $\x_1$ and $\x_2$ independently, and encode them into $\f_1 = \enc(\x_1)$ and $\f_2 = \enc(\x_2)$.
\item \textbf{Mix}: Define a mask $\m = [m^1 \mathbf{1}, m^2 \mathbf{1}, \hdots, m^n \mathbf{1}]$, where $m^i$ are uniformly sampled in $\{ 0, 1 \}$, and $\mathbf{1} = [1,1, \hdots, 1] \in \mathbf{R}^d$. Select the $i$-th feature chunk from $\f_1$ if $m^i=1$ and from $\f_2$ if $m^i=0$; collect them into a new feature $\f_{1\oplus2} = \m \odot \f_1 + (\mathbb{1}-\m) \odot f_2$, where $\odot$ is the element-wise multiplication and $\mathbb{1} = [\mathbf{1},\mathbf{1}, \hdots, \mathbf{1}]$.
\item Decode a new image $\x_3 = \dec( \f_{1\oplus2})$.
\item Encode again, $\f_3 = \enc(\x_3)$.
\item \textbf{Unmix} $\f_3$ by replacing feature chunks from $\f_2$, given by the mask $\mathbb{1}-\m$, with the corresponding ones from $\f_1$, that is, $\f_{3\oplus1} = \m \odot \f_3 + (\mathbb{1}-\m) \odot \f_1$.
\item Decode the final image $\x_4 = \dec(\f_{3\oplus1})$, from the mixed features of $\f_3$ and $\f_1$.
\end{enumerate}

Finally, we minimize the squared $L^2$ distance between  $\x_1$  and $\x_4$, thus the loss function can be written as
\begin{eqnarray}
\label{eq:Ls}
\textstyle
	{\cal L}_{M}(\theta_{\text{Enc}}, \theta_{\text{Dec}}) = 
	E_{\x_1, \x_2} \Big[  \sum_{\m} | \x_4 - \x_1 |^2 \Big],
\end{eqnarray}
where we sum over all possible mask settings, and $\theta_{\text{Enc}}$ and $\theta_{\text{Dec}}$ are the encoder and decoder parameters respectively.

\begin{figure}[t]
	\begin{subfigure}[b]{0.15\textwidth}
		\includegraphics[width=\linewidth]{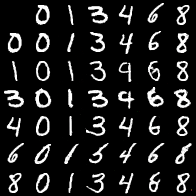} 
		\caption{Digit class}
	\end{subfigure}
	\begin{subfigure}[b]{0.15\textwidth}
		\includegraphics[width=\linewidth]{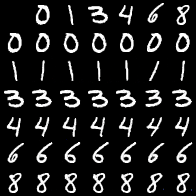} 
		\caption{Rotation angle}
	\end{subfigure}
	\begin{subfigure}[b]{0.15\textwidth}
		\includegraphics[width=\linewidth]{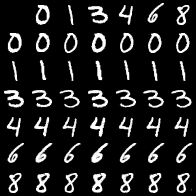} 
		\caption{Stroke width}
	\end{subfigure}
	\centering
	\vspace{-.3cm}
	\caption{Attribute transfer on the MNIST dataset by mixing individual chunks between pairs of source images, shown in the topmost row and leftmost column. 
	To generate an image in column $i$ and row $j$, we take one chunk from the $i$-th image in the top row, and the other chunks from the $j$-th image in the leftmost column. In each subfigure, the mixed chunk corresponds to the attribute indicated in the caption of the subfigure.}
	\label{fig:mnist_attributes}
\end{figure}

Intuitively, the key idea is that the cycle of decoding and re-encoding of the mixed feature vector $\f_{1\oplus2}$ should preserve the chunks from $\f_{1}$ that were copied in $\f_{1\oplus2}$. In other words, these chunks from $\f_{1}$ should be decoded into corresponding attributes of $\x_3$. 
In addition, re-encoding into $\f_3$ the intermediate image $\x_3$ consisting of a mix of attributes from $\x_1$ and attributes from $\x_2$, should return the same feature chunks originally from $\x_1$.


\begin{table}[t]
\begin{center}
	\vspace{-0.5em}
	\caption{Network architectures of encoder (\textbf{Enc}), decoder (\textbf{Dec}), discriminator (\textbf{Dsc}) and classifier (\textbf{Cls}) on different datasets. We denote the convolutional layer with ``c'', the deconvolutional layer with ``d'' and the fully connected layer with ``f''. The numbers denote the number of channels.
	The kernel size and stride are denoted with ``k'' and ``s'', and they are omitted when they are equal to $1$.
	The pooling layers ``p'' have kernel size $3$ and stride $2$.
	After each convolutional and deconvolutional layer we added a normalization and a leaky ReLU layer with a leak coefficient of $0.2$.
	For BEGAN, the discriminator architecture is the same as that of the autoencoder. We used ReLU after the convolutional layers, and ``r'' stands for reshape and ``u'' for upsampling by a factor of $2$.
	We choose $\gamma=0.5$ for training. }
	\label{table:architecture}
	\footnotesize
	\begin{tabular}{ | l | c| }
		\hline
		& \textbf{CelebA (DCGAN)} \\
		\hline
		\textbf{Enc} & c64k3s2-c128k3s2-c256k3s2-c512k3s2-c512k2-f \\
		\textbf{Dec} & d512k4-d512k4s2-d256k4s2-d128k4s2-d3k2 \\
		\textbf{Dsc} & c64k3s2-c128k3s2-c256k3s2-c512-f \\
		\textbf{Cls} & c96k8s2-p-c256k5-p-c384k3-c384k3-c256k3-p-f4096-f4096-f \\
		\hline
		\hline
		& \textbf{CelebA (BEGAN)} \\
		\hline
		\textbf{Enc} & c32k3-c32k3-c32k3-c64-p-c64k3-c64k3-c96-p-c96k3-c96k3-\\ &c128-p-c128k3-c128k3-c160-c160-p-c160k3-c160k3-f \\
		\textbf{Dec} & f4096-r(8,8,64)-c64k3-c64k3-u-c64k3-c64k3-u-c64k3-\\&c64k3-u-c64k3-c64k3-u-c64k3-c64k3-c3\\
		\textbf{Cls} & c96k8s2-p-c256k5-p-c384k3-c384k3-c256k3-p-f4096-f4096-f\\
		\hline
		\hline
		& \textbf{MNIST} \\
		\hline
		\textbf{Enc} & c64k3s2-c128k3s2-c256k3s2-f\\
		\textbf{Dec} & d512k4-d256k4s2-d128k4s2-d3k2\\
		\textbf{Dsc} & c64k3s2-c128k3s2-c256k3s2-c512-f \\
		\textbf{Cls} & c96k8s2-p-c256k5-p-c384k3-c384k3-c256k3-p-f4096-f4096-f \\
		\hline
		\hline
		& \textbf{Sprites} \\
		\hline
		\textbf{Enc} & c64k3s2-c128k3s2-c256k3s2-c512k2s2-c512k2-f \\
		\textbf{Dec} & d512k4-d512k4s2-d256k4s2-d128k4s2-d3k2\\
		\textbf{Dsc} & c64k3s2-c128k3s2-c256k3s2-c512-f \\
		\textbf{Cls} & c96k8s2-p-c256k5-p-c384k3-c384k3-c256k3-p-f4096-f4096-f \\
		\hline
	\end{tabular}
	\vspace{-2em}
\end{center}
\end{table}

\noindent\textbf{Discriminator.} To ensure that the generated perturbed images $\x_3$ are valid images according to the input data distribution, we impose an additional adversarial term, which is defined as 
\begin{align}
\label{eq:Lg}
	{\cal L}_{G}(\theta_{\text{Enc}}, \theta_{\text{Dec}}, \theta_{\text{Dsc}}) = &\\
\textstyle
	\sum_{\m} E_{\x_1,\x_2} \Big[ & \log( \dsc (\x_1)) + \log(1- \dsc( \x_3 ))  \Big],\nonumber
\end{align}
where $\theta_{\text{Dsc}}$ are the discriminator parameters. In the ideal case when the GAN objective reaches the global optimum, the distribution of fake images should match the real image distribution. 
With the invariance and adversarial loss, however, is still possible to encode all image attributes into one feature chunk and keep the rest constant. This solution optimizes both the invariance loss and the adversarial loss perfectly. 
As mentioned before, this is called the shortcut problem and we address it using an additional loss based on a classification task.

\begin{figure*}[t]
\centering
	\begin{subfigure}[b]{\spritessize\textwidth}
		\includegraphics[width=\linewidth]{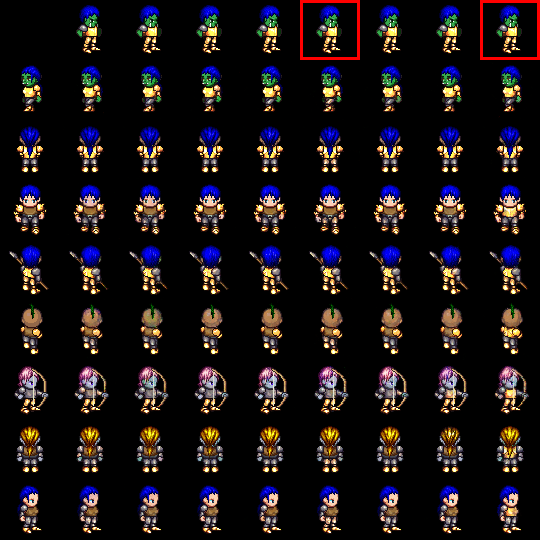} 
		\caption{MIX}
	\end{subfigure}
	\begin{subfigure}[b]{\spritessize\textwidth}
		\includegraphics[width=\linewidth]{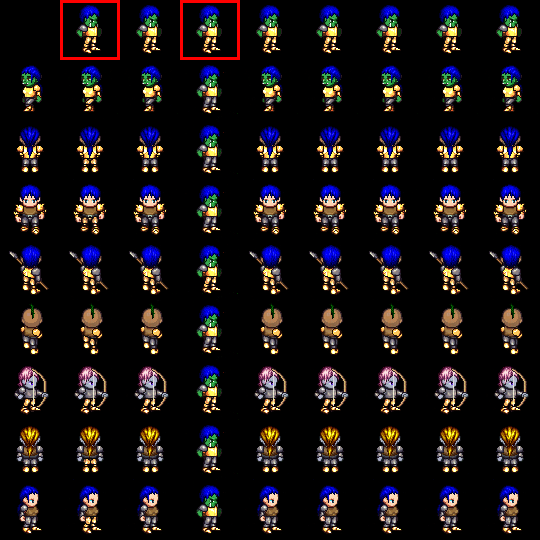} 
		\caption{MIX+G}
	\end{subfigure}
	\begin{subfigure}[b]{\spritessize\textwidth}
		\includegraphics[width=\linewidth]{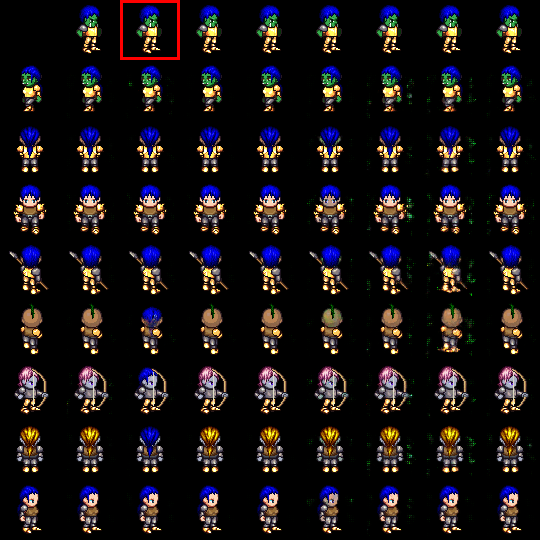} 
		\caption{MIX+C}
	\end{subfigure}
	\begin{subfigure}[b]{\spritessize\textwidth}
		\includegraphics[width=\linewidth]{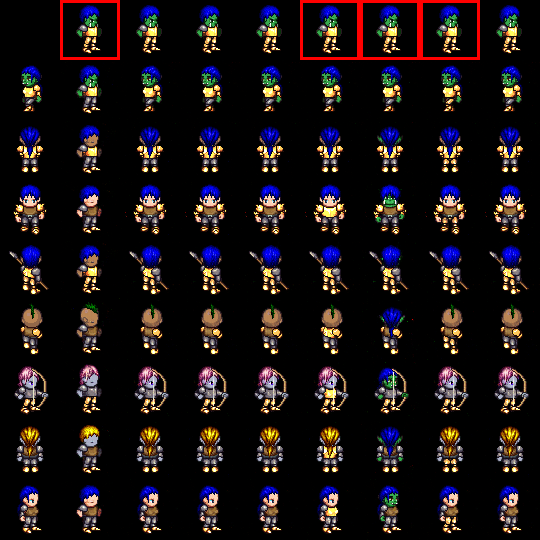} 
		\caption{MIX+G+C}
	\end{subfigure}
	\vspace{-.3cm}
	\caption{Comparison of different methods on Sprites. In all subfigures, images are generated by taking one of the $8$ feature chunks from the topmost row, and the others from the leftmost column. Red frames indicate whether a feature chunk encodes an attribute.
	MIX denotes the mixing loss, G the adversarial loss and C the classifier loss in the objective.
	MIX+G+C disentangles  pose, torso color, hair color, and leg color (columns marked with red boxes, left to right).}
	\label{fig:sprites_methods}
\end{figure*}

\noindent\textbf{Classifier.} The last component of our network takes three images as inputs: the input images $\x_1$ and $\x_2$, and the generated image $\x_3$. It decides for every chunk whether the composite image was generated using the feature from the first or the second input image. The formal loss function is
\begin{align}
\label{eq:Lc}
	{\cal L}_{C}(\theta_{\text{Enc}}, \theta_{\text{Dec}}, \theta_{\text{Cls}}) = &\\
	\textstyle
E_{\x_1,\x_2} \Big[ - \sum_{\m} \sum_i & m^i \log(y^i) + (1-m^i) \log(y^i)) \Big],\nonumber
\end{align}
where $\theta_{\text{Cls}}$ are the classifier parameters, and its outputs are $\cls(\x_1, \x_2, \x_3) = \y = [y^1, y^2, \hdots, y^n]$. The classifier consists of $n$ binary classifiers, one for each chunk, that decide whether the composite image $\x_3$ was generated using the corresponding chunk from the first image or the second. We use the cross entropy loss for classification, so the last layer of the classifier is a sigmoid. The classifier loss can only be minimized if there is a meaningful attribute encoded in every chunk. Hence, the shortcut problem cannot occur as it would be impossible to decide which chunks were used to create the composite image.

Finally, our overall objective consists of the weighted sum of the three components described above, 
\begin{eqnarray}
\textstyle
	\min_{\theta_{\text{Enc}}, \theta_{\text{Dec}}, \theta_{\text{Cls}}} \max_{\theta_{\text{Dsc}}} \quad
	\lambda_{M} {\cal L}_{M} + \lambda_{G} {\cal L}_{G} + \lambda_{C} {\cal L}_{C}.
\end{eqnarray}
Note that during training, we randomly sample the masks $\m$ instead of computing a sum over all possibilities for all image sample pairs (Eqns.~\eqref{eq:Ls},~\eqref{eq:Lg} and~\eqref{eq:Lc}).




\begin{figure}
	\begin{center}
		\includegraphics[width=0.6\linewidth,trim={2cm 1.5cm 2cm 0.7cm}]{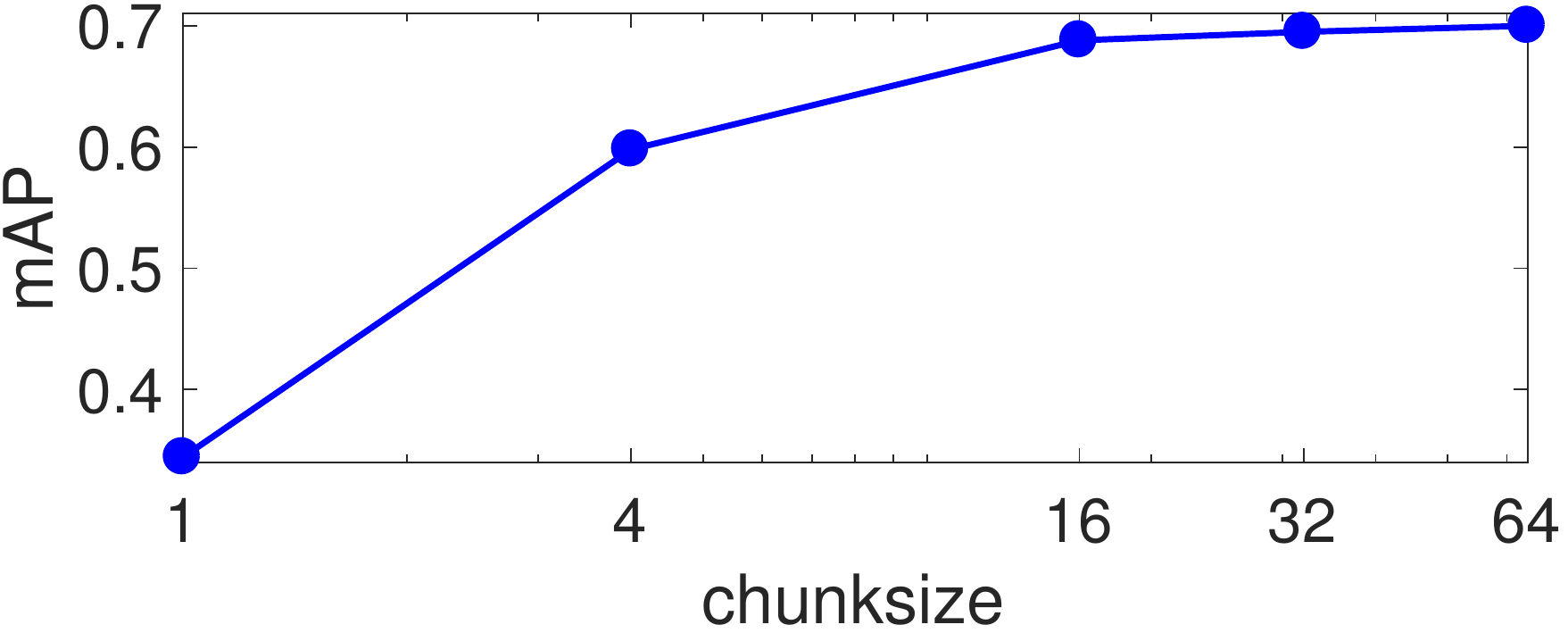}
	\end{center}
	\caption{Mean average precision of the nearest neighbor search averaged across labeled attributes, as a function of the chunk size on the Sprites dataset with the complete model: the mixing autoencoder + the classifier + GAN.}
	\label{fig:chunksize}
\end{figure}

\subsection{Implementation}
We use a network architecture similar to DCGAN~\cite{dcgan2015} for the encoder, decoder, and discriminator. For the classifier, we use AlexNet with batch normalization after each convolutional layer, but we do not use any dropout. The image inputs of the classifier are concatenated along the RGB channels. We use equal weights $\lambda_{M} = \lambda_{G}=\lambda_{C}=1$ for the mixing autoencoder, GAN, and classifier for our experiments on the MNIST and Sprites datasets. For CelebA, we increase the weight of the mixing autoencoder to $\lambda_{M}=30$. In all experiments, the feature vector is the output of the last layer of the encoder. We separate it into $8$ chunks, where each chunk is expected to represent one attribute, with equal size for each of the eight chunks.
Our results are obtained with chunk size $8$ for MNIST, $64$ for Sprites and $64$ for CelebA. We observed that reducing the chunk size in CelebA leads to lower rendering quality. For CelebA, we also show experiments using BEGAN~\cite{began2017} for the adversarial training. The detailed architectures are shown in Table~\ref{table:architecture}.


\section{Experiments}

We experimented on three
public datasets, the MNIST handwritten digits \cite{MNIST}, Sprites animated figures \cite{reed2015deep}, and CelebA faces \cite{CelebA}.
We show qualitative results on all datasets and quantitative evaluations and ablation studies on Sprites and CelebA.
~\\\noindent\textbf{MNIST.}
The MNIST dataset consists of $60$K handwritten digits for the training and $10$K for the test set, given as grayscale images with a size of $28\times28$ pixels. There are $10$ different classes referring to the different digits. Other attributes like rotation angle or stroke width are not labeled. Our method can disentangle the labeled attribute as well as some non-labeled ones. Figure~\ref{fig:mnist_attributes} shows visual attribute transfers for three factors: digit class, rotation angle, and stroke width. The three chunks were chosen by visually inspecting which chunk corresponded to which attribute.
All discernible variations seem to be encoded in the three chunks, and transferring the other chunks seem to have little visual effect.
\begin{table*}[t]
	\caption{
	Mean average precision performance of nearest neighbor classification on the Sprites dataset, which comes with labeled attributes. Each row contains different methods, while the columns show the classification performance of different attributes.
	MIX denotes the mixing loss, G the adversarial loss, C the classifier loss and AE is the vanilla autoencoder in the objective.}
\vspace{-0.5em}
\centering
\small
	\begin{tabular}{ | l | c c c c c c c | c |}
		\hline
		Method & body & skin & vest & hair & arm & leg & pose & average\\
		\hline
		Random & 0.5 & 0.25 & 0.33 & 0.17 & 0.5 & 0.5 & 0.006 & 0.32 \\
		C+G & 0.53 & 0.31 & 0.41 & 0.24 & 0.51 & 0.52 & 0.06 & 0.37 \\
		AE & 0.56 & 0.37 & 0.40 & 0.31 & 0.54 & 0.56 & 0.46 & 0.46 \\
		AE+C+G & 0.59 & 0.50 & 0.53 & 0.46 & 0.56 & 0.54 & 0.44 & 0.52 \\
		MIX & 0.57 & 0.61 & 0.51 & 0.62 & 0.54 & 0.94 & \textbf{0.53} & 0.62 \\
		MIX + C & 0.57 & 0.65 & 0.43 & \textbf{0.63} & 0.55 & 0.58 & 0.51 & 0.56 \\
		MIX + G & \textbf{0.59} & 0.31 & 0.44 & 0.24 & 0.54 & 0.96 & 0.47 & 0.51 \\
		MIX + C + G & 0.58 & \textbf{0.80} & \textbf{0.94} & 0.49 & \textbf{0.58} & \textbf{0.96} & 0.52 & \textbf{0.70} \\
		\hline
	\end{tabular}
	\label{spritestable}
\end{table*}
\begin{figure*}[t]
	\center
	\begin{subfigure}[b]{\spritessize\textwidth}
		\stackunder[1pt]{\includegraphics[width=\linewidth]{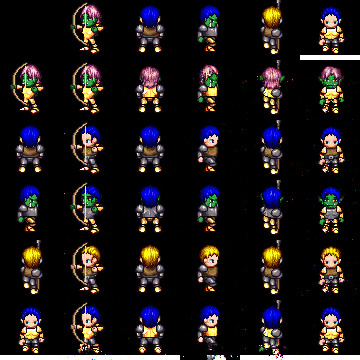}}{(a) Pose+arm}
	\end{subfigure}
	\begin{subfigure}[b]{\spritessize\textwidth}
		\stackunder[1pt]{\includegraphics[width=\linewidth]{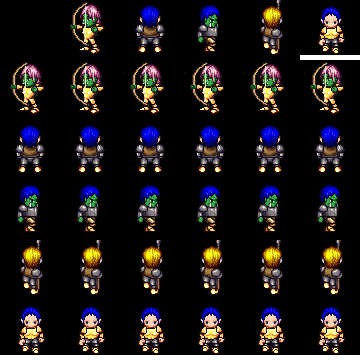}}{(b) Undefined}
	\end{subfigure}
	\begin{subfigure}[b]{\spritessize\textwidth}
		\stackunder[1pt]{\includegraphics[width=\linewidth,]{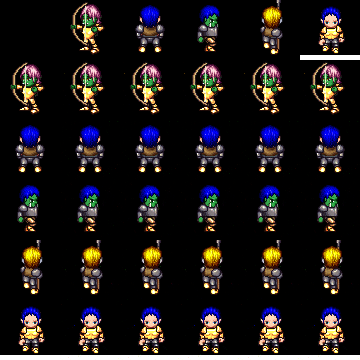}}{(c) Undefined}
	\end{subfigure}
	\begin{subfigure}[b]{\spritessize\textwidth}
		\stackunder[1pt]{\includegraphics[width=\linewidth]{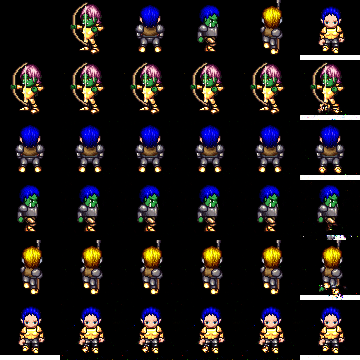}}{(d) White bar}
	\end{subfigure}
	
	\begin{subfigure}[b]{\spritessize\textwidth}
		\stackunder[1pt]{\includegraphics[width=\linewidth,trim=0  0 0 0]{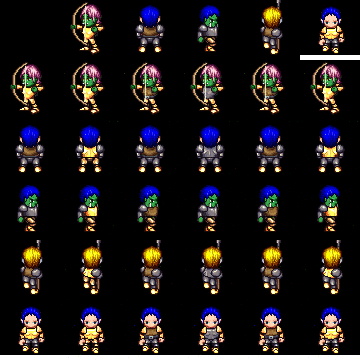}}{(e) Vest} 
	\end{subfigure}
	\begin{subfigure}[b]{\spritessize\textwidth}
		\stackunder[1pt]{\includegraphics[width=\linewidth,trim=0  0 0 0]{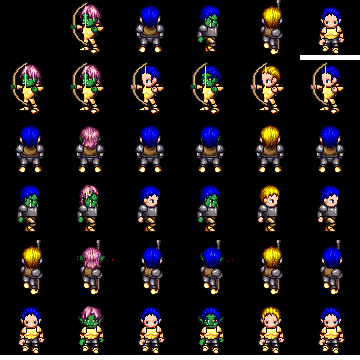}}{(f) Skin+hair}
	\end{subfigure}
	\begin{subfigure}[b]{\spritessize\textwidth}
		\stackunder[1pt]{\includegraphics[width=\linewidth,trim=0  0 0 0]{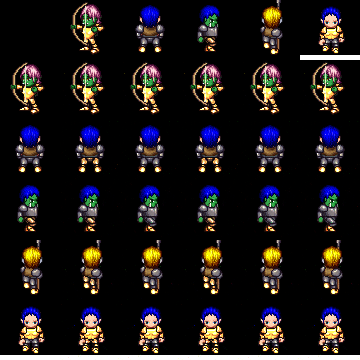}}{(g) Leg} 
	\end{subfigure}
	\begin{subfigure}[b]{\spritessize\textwidth}
		\stackunder[1pt]{\includegraphics[width=\linewidth]{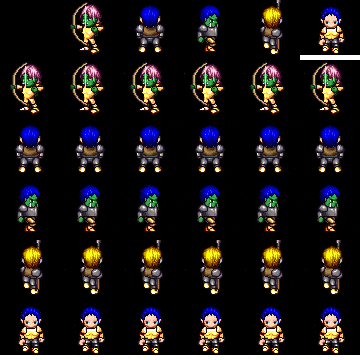}}{(h) Undefined}
	\end{subfigure}
	
	\centering
	\caption{Attribute transfer on the Sprites dataset. For every subfigure (a) to (h), one of the eight chunks is taken from the topmost row and the rest from the leftmost column. Each subfigure visualizes the role of one of the eight chunks, and the subfigure captions indicate the attribute (if semantically meaningful) associated with the chunk.}
	\label{fig:sprites_attributes}
	\vspace{-1em}
\end{figure*}
~\\\noindent\textbf{Sprites.}
The sprites dataset has $672$ synthetically rendered animated characters (sprites). The dataset is split into a training set with $500$, a validation set with $72$, and a test set with $100$ sprites. Each sprite is rendered at $178$ positions, thus the number of images is $120$K in total. The dataset has many labeled attributes: body shape, skin color, vest color, hairstyle, arm and leg color, and finally weapon type. The pose labels can be extracted from the frame number of the animations. This rich attribute labeling is ideal for testing the disentanglement of our algorithms.

We perform ablation studies on the components of our method.
The qualitative results are shown in Figure~\ref{fig:sprites_methods}.
We can see that mixing autoencoder already learned to disentangle $2$ chunks. Adding only GAN does not improve the disentangling, as its job is to make the images look more realistic. However, the rendering quality without GAN is already good.
Adding only the classifier does not improve disentangling either, it rather creates artifacts in the rendering.
The intuitive explanation is that the classifier solves the shortcut problem in the sense that it forces all chunks to carry information about the inputs. However, the information seem to be stored as artifacts, while the interpretable attributes are ignored.
The full objective with all three components on the other hand improves the performance, as the artifacts are eliminated by GAN, and the shortcut problem can only be avoided by disentangling the factors. The method recovers $4$ independent factors.
\begin{figure*}[t]
\centering
	\rotatebox{90}{\hspace{3.5cm} \small (a) DCGAN}
\begin{subfigure}[b]{.95\textwidth}
	\centering
	\begin{subfigure}[b]{\celebasize\textwidth}
	\stackunder[1pt]
		{\includegraphics[width=\linewidth]{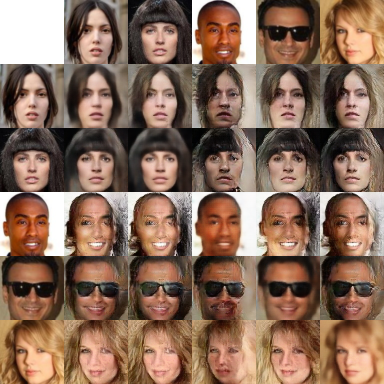}}{Undefined}
		\label{fig:celebA_a} 
	\end{subfigure}
	\begin{subfigure}[b]{\celebasize\textwidth}
	\stackunder[1pt]
		{\includegraphics[width=\linewidth]{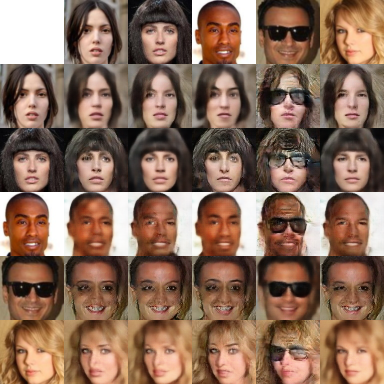}}{Glasses}
		\label{fig:celebA_b} 
	\end{subfigure}
	\begin{subfigure}[b]{\celebasize\textwidth}
	\stackunder[1pt]
		{\includegraphics[width=\linewidth,]{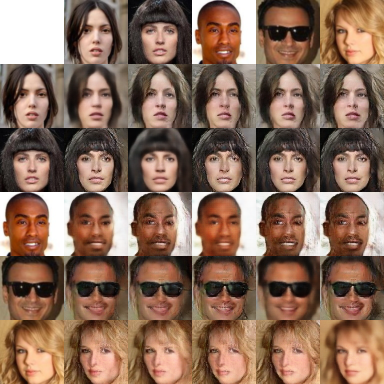}}{Undefined}
		\label{fig:celebA_c} 
	\end{subfigure}
	\begin{subfigure}[b]{\celebasize\textwidth}
	\stackunder[1pt]
		{\includegraphics[width=\linewidth,trim=0  0 0 0]{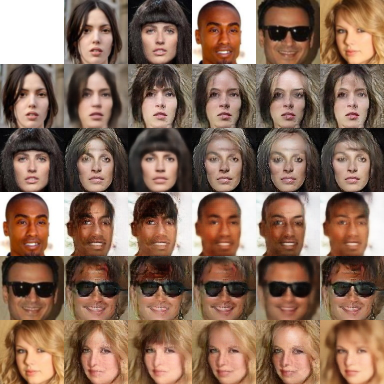}}{Hair style}
		\label{fig:celebA_d} 
	\end{subfigure}
	
	\begin{subfigure}[b]{\celebasize\textwidth}
	\stackunder[1pt]
		{\includegraphics[width=\linewidth,trim=0  0 0 0]{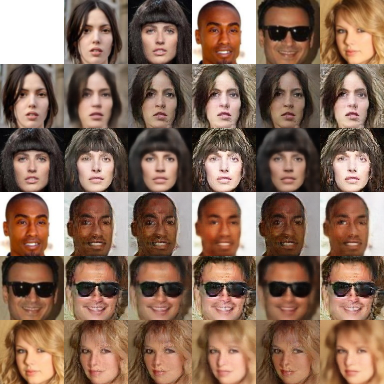} }{Brightness}
		\label{fig:celebA_e} 
	 \end{subfigure}
	 \begin{subfigure}[b]{\celebasize\textwidth}
	 \stackunder[1pt]
	 	{\includegraphics[width=\linewidth,trim=0  0 0 0]{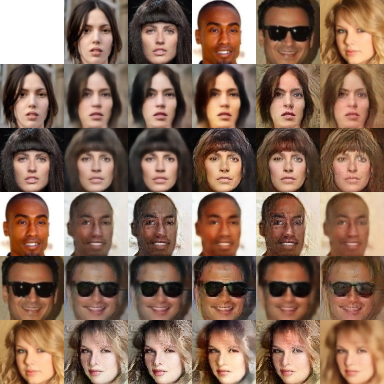}}{Hair color} 
		\label{fig:celebA_f} 
	\end{subfigure}
	\begin{subfigure}[b]{\celebasize\textwidth}
	\stackunder[1pt]
		{\includegraphics[width=\linewidth,trim=0  0 0 0]{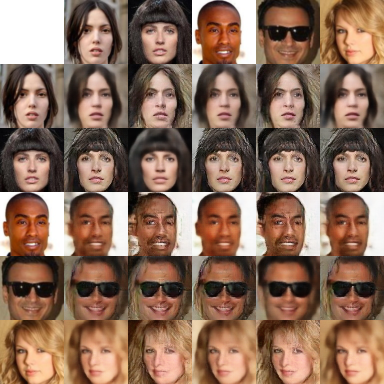} }{Undefined}
		\label{fig:celebA_g} 
	\end{subfigure}
	\begin{subfigure}[b]{\celebasize\textwidth}
	\stackunder[1pt]
		{\includegraphics[width=\linewidth]{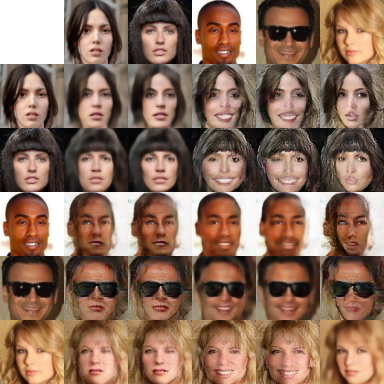} }{Pose/smile}
	\label{fig:celebA_h} 
	\end{subfigure}
\end{subfigure}\\
	\rotatebox{90}{\hspace{3.5cm} \small (b) BEGAN}
\begin{subfigure}[b]{.95\textwidth}
	\centering
	\begin{subfigure}[b]{\celebasize\textwidth}
	\stackunder[1pt]
		{\includegraphics[width=\linewidth]{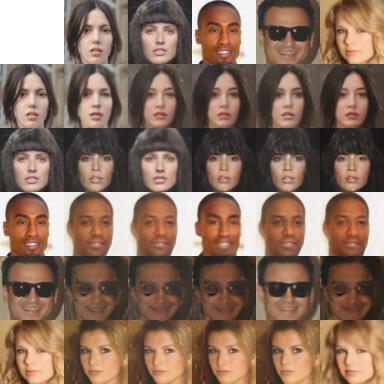}}{Undefined}
		\label{fig:celebA_a} 
	\end{subfigure}
	\begin{subfigure}[b]{\celebasize\textwidth}
	\stackunder[1pt]
		{\includegraphics[width=\linewidth]{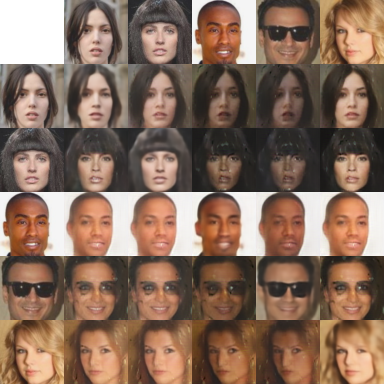}}{Brightness}
		\label{fig:celebA_b} 
	\end{subfigure}
	\begin{subfigure}[b]{\celebasize\textwidth}
	\stackunder[1pt]
		{\includegraphics[width=\linewidth,]{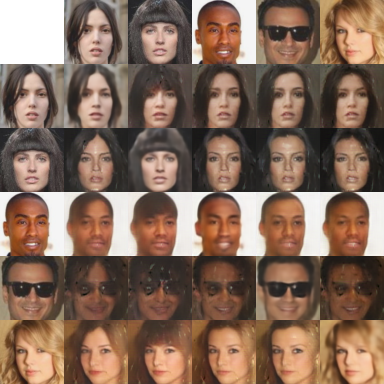}}{Hair style} 
		\label{fig:celebA_c} 
	\end{subfigure}
	\begin{subfigure}[b]{\celebasize\textwidth}
	\stackunder[1pt]
		{\includegraphics[width=\linewidth]{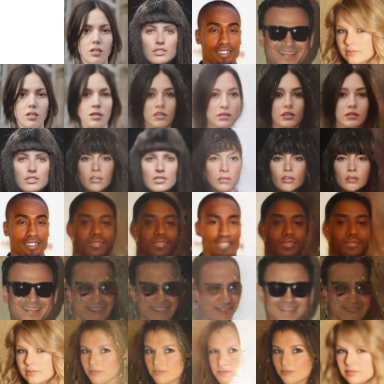}}{Background}
		\label{fig:celebA_a} 
	\end{subfigure}

	\begin{subfigure}[b]{\celebasize\textwidth}
	\stackunder[1pt]
		{\includegraphics[width=\linewidth,trim=0  0 0 0]{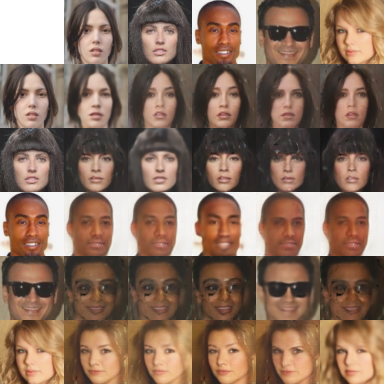}}{Glasses}
		\label{fig:celebA_a} 
	 \end{subfigure}
	 \begin{subfigure}[b]{\celebasize\textwidth}
	 \stackunder[1pt]
	 	{\includegraphics[width=\linewidth,trim=0  0 0 0]{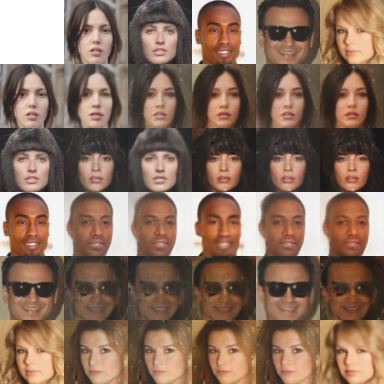}}{Undefined}
		\label{fig:celebA_a} 
	\end{subfigure}
	\begin{subfigure}[b]{\celebasize\textwidth}
	\stackunder[1pt]
		{\includegraphics[width=\linewidth,trim=0  0 0 0]{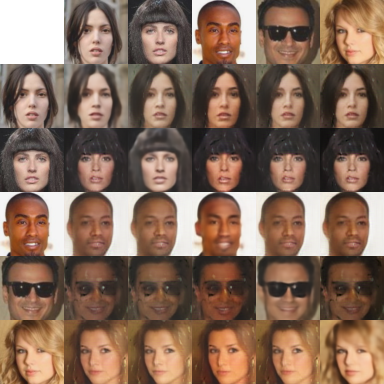}}{Saturation} 
		\label{fig:celebA_a} 
	\end{subfigure}
	\begin{subfigure}[b]{\celebasize\textwidth}
	\stackunder[1pt]
		{\includegraphics[width=\linewidth]{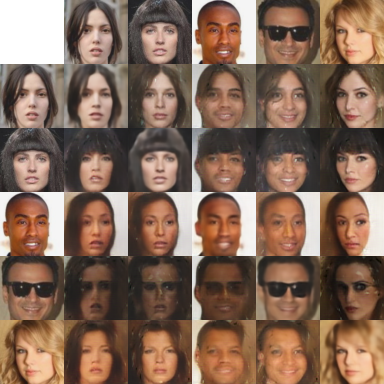}}{Pose/gender}
	\label{fig:celebA_began_a} 
	\end{subfigure}
\end{subfigure}
\centering
\caption{Attribute transfer on the CelebA dataset with our method using (a) DCGAN and (b) BEGAN. For every subfigure, one chunk is taken from the topmost row and the rest from the leftmost column. Different subfigures show the role of different chunks. The captions indicate the attribute associated with the chunk.}
\label{fig:celebA_attributes}
\vspace{-1.5em}
\end{figure*}

For quantitative analysis we perform nearest neighbor search using a chunk of the features and compute the mean average precision using an attribute as ground truth. We repeat the search for all chunk and attribute pairs, and for each attribute we choose the best performing chunk to represent it.
We ignore the weapon type attribute in our evaluation, as it is only visible in a small subset of poses.
We also compare our method to the vanilla autoencoder. It has only one chunk, but its dimensionality is the same as the full feature of the other methods.
In Table~\ref{spritestable} we compare the results of our methods.
We can see a consistent improvement of our proposed mixing autoencoder over the vanilla autoencoder, whether we use the classifier and the GAN or not. The classifier and the GAN together also consistently help, no matter which autoencoder was used (MIX, vanilla or none). The classifier or the GAN alone do not help the performance, which is in line with the qualitative experiments as well.
\begin{table*}[t]
\small
	\caption{The classification performance on CelebA. Each row contains different methods, while the columns show the different attributes (``eyebr.'' is arched eyebrows and ``attr.'' is attractive).
	}
\vspace{-0.5em}
\centering
	\begin{tabular}{ | l | c c c c c c c c c c c c | c |}
		\hline
		Method & Eyebr. & Attr. & Bangs & Black & Blond & Makeup & Male & Mouth & Beard & Wavy & Hat & Lips & Avg. \\
		\hline\hline
		VAE & 71.8 & 73.0 & 89.8 & 78.0 & 88.9 & 79.6 & 83.9 & 76.3 & 87.3 & 70.2 & 95.8 & 83.0 & 81.5 \\
		$\beta$=2 & 71.6 & 72.6 & 90.6 & 79.3 & 89.1 & 79.3 & 83.5 & 76.1 & 86.9 & 67.8 & 95.9 & 82.4 & 81.3 \\
		$\beta$=4 & 71.6 & 72.6 & 90.0 & 76.6 & 88.9 & 77.8 & 82.3 & 75.7 & 85.3 & 66.8 & 95.8 & 80.6 & 80.3 \\
		$\beta$=8 & 71.6 & 71.7 & 90.0 & 76.0 & 87.2 & 76.2 & 80.5 & 73.1 & 85.3 & 63.7 & 95.8 & 79.6 & 79.2 \\
		DIP-VAE & \textbf{73.7} & \textbf{73.2} & \textbf{90.9} & \textbf{80.6} & \textbf{91.9} & \textbf{81.5} & \textbf{85.9} & 75.9 & 85.3 & \textbf{71.5} & 96.2 & \textbf{84.7} & \textbf{82.6} \\
		\hline
		Ours (DCGAN) & 72.2 & 68.5 & 88.8 & 75.7 & 89.9 & 76.9 & 80.1 & 73.6 & 83.8 & 70.5 & 95.8 & 78.6 & 79.5 \\
		Ours (BEGAN) & 73 & 69.7 & 90.2 & 79.6 & 89.3 & 78.9 & 85.4 & \textbf{77.1} & \textbf{88.1} & 70.8 & \textbf{96.4} & 81.7 & 81.7 \\
		\hline
	\end{tabular}
	\label{celebatable}
\end{table*}
\begin{figure*}[!t]
	\centering
	\begin{subfigure}[b]{0.19\textwidth}
		\includegraphics[width=\linewidth]{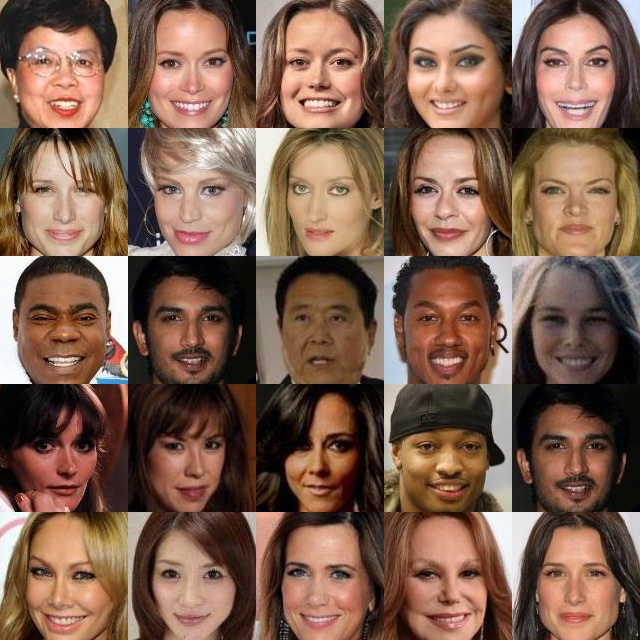} 
		\caption{\vspace{-.2cm}Brightness}
		\label{fig:face_color} 
	\end{subfigure}
	\begin{subfigure}[b]{0.19\textwidth}
		\includegraphics[width=\linewidth]{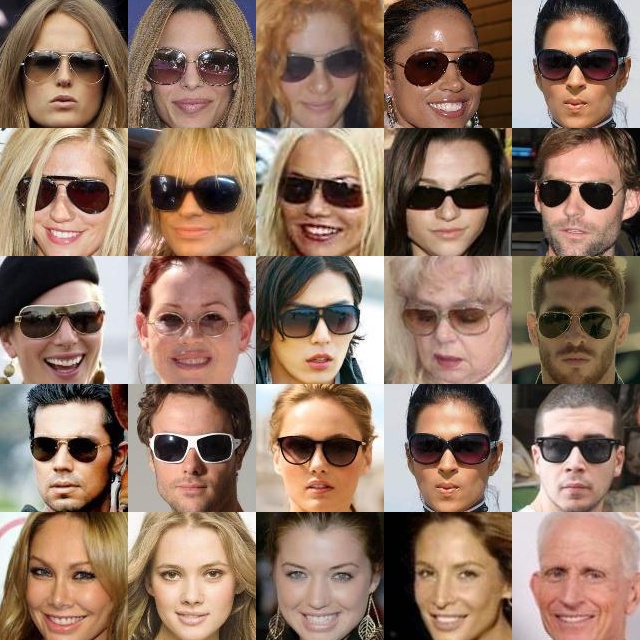} 
		\caption{\vspace{-.2cm}Glasses}
		\label{fig:glasses}
	\end{subfigure}
	\begin{subfigure}[b]{0.19\textwidth}
		\includegraphics[width=\linewidth,]{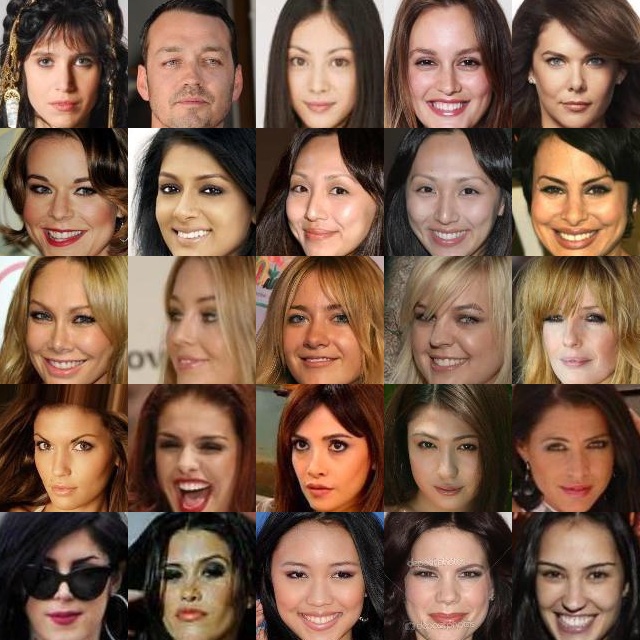} 
		\caption{\vspace{-.2cm}Hair color}
		\label{fig:haircolor} 
	\end{subfigure}
	\begin{subfigure}[b]{0.19\textwidth}
		\includegraphics[width=\linewidth]{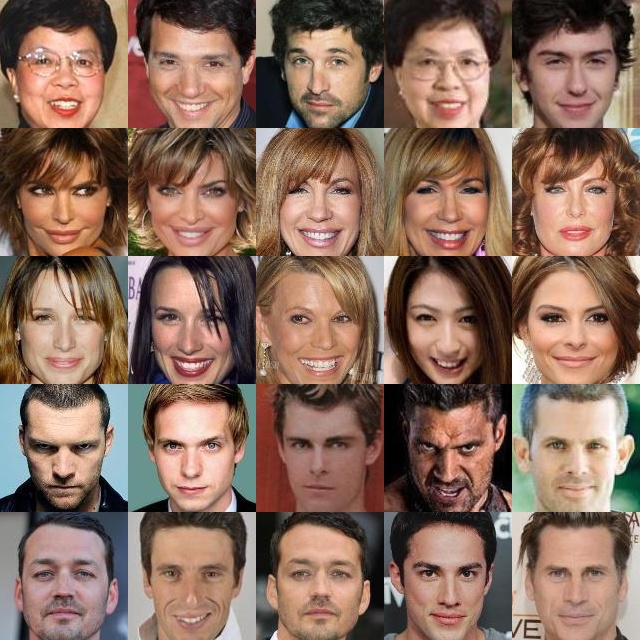} 
		\caption{\vspace{-.2cm}Hair style}
		\label{fig:hairstyle} 
	\end{subfigure}
	\begin{subfigure}[b]{0.19\textwidth}
		\includegraphics[width=\linewidth,trim=0  0 0 0]{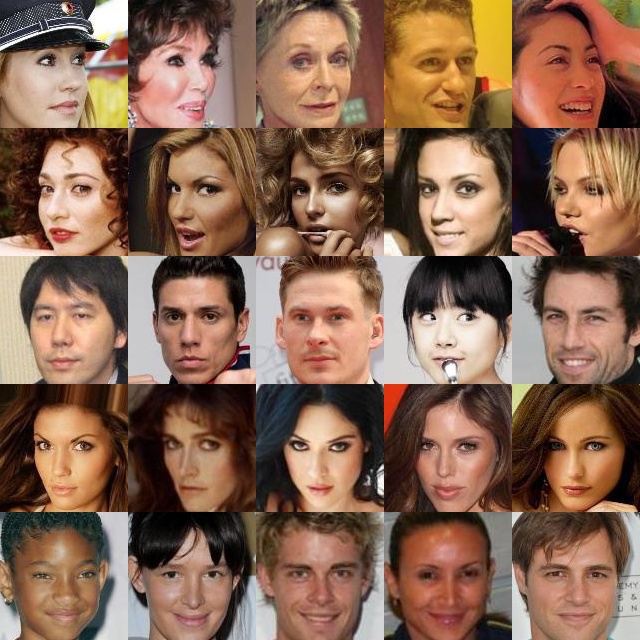} 
		\caption{\vspace{-.2cm}Pose/smile}
		\label{fig:posesmile}
	\end{subfigure}
	\caption{Image retrieval on CelebA of our method with DCGAN. Subfigures show the nearest neighbor matches for different feature chunks. For all subfigures, the first column contains the query images and subsequent columns contain the top matches using the $L^2$ distance. The caption indicates the discovered semantic meaning.}
	\label{fig:celebA_nearest}
\end{figure*}
Figure~\ref{fig:chunksize} shows the effect of the chunk size on the classification performance. Increasing the number of dimensions helps, but we reach a plateau at $16$ dimensions. We chose a large chunk size $64$ for our experiments to better highlight that we can avoid the shortcut problem, the degenerate solution where all information is stored in one chunk.
Figure~\ref{fig:sprites_attributes} visualizes attribute transfer for all chunks, similarly to Figure~\ref{fig:mnist_attributes}, using our complete method. We can recover the leg and vest colors into single chunks, while the pose and arm color attribute pair is represented by one chunk. The skin color and hairstyle attributes are also entangled and represented by another chunk. There are $6$ positions, where the sprites stand on a white bar. Even though this attribute is fully determined by the position, our method separates it to its own chunk.
~\\\noindent\textbf{CelebA.}
CelebA contains $200$K color images of celebrity faces. The training, validation, and test sizes are $160$K, $20$K and $20$K respectively. There are $40$ labeled binary attributes indicating gender, hair color, facial hair and so on. We applied our method with both BEGAN and DCGAN architectures.
Figure~\ref{fig:celebA_attributes} shows the attribute transfer for each chunk. We can see that DCGAN exhibits more pronounced attribute transfer, while BEGAN tends to blur out the changes.
Figure~\ref{fig:celebA_nearest} shows the nearest neighbors of some query images in the dataset using DCGAN. We used the $L^2$ distance on the specified feature chunks to search for top matches. For each chunk the top matches preserve a semantic attribute of the query image. Our method could recover five semantically meaningful attributes: brightness, glasses, hair color, hair style, and pose and smile. Notice that the attributes discovered with attribute transfer match the attributes in image retrieval. For brevity we only show those five chunks.
We performed quantitative tests on our learned features. We followed the evaluation technique based on the equivariant disentanglement property described in \cite{dipvae2017}. A feature representation is considered disentangled when the attributes can be classified using a simple linear classifier. In our special case when an attribute depends only on one chunk (a subspace), a linear classifier would perform well by setting the classifier weights with respect to the other chunks to zero. We train binary classifiers on the whole feature vector, each with a different labeled attribute as ground truth. The classifier prediction is $\text{sign}(\w^T \f + b)$, where the classifier weights are computed as
\begin{align}
	\textstyle \w = \frac{1}{ | i : c_i = +1 | } \sum_{i : c_i = +1} \f_i - \frac{1}{ | i : c_i = -1 | } \sum_{i : c_i = -1} \f_i,
\end{align}
where $c_i \in \{-1,+1\}$ are the attribute labels. The bias term $b$ is set by minimizing the hinge loss.
For a fair comparison we normalize the features by setting the variance for each coordinate to one, as in \cite{dipvae2017} the features are already normalized by the variational autoencoder.
The results are shown in Table~\ref{celebatable}. 
We can see that our network is competitive with the state of the art methods, $\beta$-VAE~\cite{higgins2016beta} and concurrent work DIP-VAE~\cite{dipvae2017}. The BEGAN architecture performs slightly better than DCGAN, despite the superior rendering quality of the latter.

\section{Conclusions}

We have introduced a novel method to disentangle factors of variation of a single set of images where no annotation is available. Our representation is computed through an autoencoder, which is trained by imposing constraints between the encoded features and the rendered images. We train the decoder to render realistic images by feeding features obtained by randomly mixing features from two images and by using adversarial training. Moreover, we force the autoencoder to make full use of the features by training it jointly with a classifier that determines how features have been mixed from an input image. We show that this technique successfully disentangles factors of variation in the MNIST, Sprites and CelebA datasets.

\noindent\textbf{Acknowledgements.} QH, TP and AS have been supported by the Swiss National Science Foundation
(SNSF) grants 200021\_149227 and 200021\_156253.


{\small
\bibliographystyle{ieee}
\bibliography{refs}
}

\end{document}